\documentclass{article}

\usepackage{times}
\usepackage{natbib}

\usepackage{hyperref}

\usepackage[accepted]{icml2017}

\usepackage{times}

\usepackage{amssymb, amsmath}
\usepackage{mathtools}
\usepackage{graphicx}
\usepackage{subfigure}
\usepackage{enumitem}
\usepackage{url}            % simple URL typesetting
\usepackage{booktabs}       % professional-quality tables
\usepackage{amsfonts}       % blackboard math symbols
\usepackage{nicefrac}       % compact symbols for 1/2, etc.
\usepackage{microtype}      % microtypography
\usepackage{color}

\makeatletter
\def\BSTATE{\STATE \hskip-3mm}
\makeatother

\newcommand*{\x}{\mathbf{x}}
\newcommand*{\xk}{\mathbf{x}_k}
\newcommand*{\xkk}{\mathbf{x}_{k+1}}
\newcommand*{\uu}{\mathbf{u}}
\newcommand*{\uk}{\mathbf{u}_k}
\newcommand*{\huk}{\hat{\mathbf{u}}_k}
\newcommand*{\dt}{\Delta t}
\newcommand*{\qk}{q_k\dt}
\newcommand*{\E}{\mathbb{E} }
\newcommand*{\xx}{\x_{k+1}|\xk }
\newcommand*{\xu}{\xk,\uk }
\newcommand*{\xxu}{\x_{k+1}|\xu }

\newcommand*{\Vavg}{V_{avg} }
\newcommand*{\hVavg}{\hat{V}_{avg}}

\newcommand*{\expq}{e^{-q_k\dt} }

\newcommand*{\R}{S}
\newcommand*{\hR}{\hat{S}}
\newcommand*{\Rinv}{\R^{-1} }

\newcommand*{\Zk}{Z_k }
\newcommand*{\Zkk}{Z_{k+1} }
\newcommand*{\Zkt}{\hat{Z}_{k} }
\newcommand*{\Zkkt}{\hat{Z}_{k+1} }
\newcommand*{\Zavg}{Z_{avg}}
\newcommand*{\hZavg}{\hat{Z}_{avg}}

\newcommand*{\paramz}{\boldsymbol{\nu} }

\newcommand*{\hdVk}{\dot{\hat{V}}_{k}}

\newcommand*{\Vkt}{\hat{V}_{k} }

\newcommand*{\trim}[1]{}

\newcommand*{\comm}[1]{}
\def\diag{\mathop{\rm diag}\nolimits}

\icmltitlerunning{Freeway Merging in Congested Traffic based on Multipolicy Decision Making with Passive Actor Critic}

\begin{document}

\twocolumn[
\icmltitle{Freeway Merging in Congested Traffic based on Multipolicy Decision Making with Passive Actor Critic}

% It is OKAY to include author information, even for blind
% submissions: the style file will automatically remove it for you
% unless you've provided the [accepted] option to the icml2017
% package.

% list of affiliations. the first argument should be a (short)
% identifier you will use later to specify author affiliations
% Academic affiliations should list Department, University, City, Region, Country
% Industry affiliations should list Company, City, Region, Country

% you can specify symbols, otherwise they are numbered in order
% ideally, you should not use this facility. affiliations will be numbered
% in order of appearance and this is the preferred way.
\icmlsetsymbol{co}{*}

\begin{icmlauthorlist}
\icmlauthor{Tomoki Nishi}{co,tcrdl,trina}
\icmlauthor{Prashant Doshi}{uog}
\icmlauthor{Danil Prokhorov}{trina}
\end{icmlauthorlist}

\icmlaffiliation{tcrdl}{Toyota Central R \& D Labs., Inc., Nagakute, Aichi, Japan}
\icmlaffiliation{uog}{THINC Lab, Dept. of Computer Science, University of Georgia, Athens, GA 30622}
\icmlaffiliation{trina}{Toyota R \& D, Ann Arbor, MI 48105}

\icmlcorrespondingauthor{Tomoki Nishi}{nishi@mosk.tytlabs.co.jp}

% You may provide any keywords that you
% find helpful for describing your paper; these are used to populate
% the "keywords" metadata in the PDF but will not be shown in the document
\icmlkeywords{autonomous vehicle, reinforcement learning, linearly solvable MDPs}

\vskip 0.3in
]

% this must go after the closing bracket ] following \twocolumn[ ...

% This command actually creates the footnote in the first column
% listing the affiliations and the copyright notice.
% The command takes one argument, which is text to display at the start of the footnote.
% The \icmlEqualContribution command is standard text for equal contribution.
% Remove it (just {}) if you do not need this facility.

\printAffiliationsAndNotice{}  % leave blank if no need to mention equal contribution
%\printAffiliationsAndNotice{\icmlEqualContribution} % otherwise use the standard text.
%\footnotetext{hi}

\begin{abstract}
Freeway merging in congested traffic is a significant challenge toward fully automated driving. Merging vehicles need to decide not only how to merge into a spot, but also where to merge. We present a method for the freeway merging based on multi-policy decision making with a reinforcement learning method called {\em passive actor-critic} (pAC), which learns with less knowledge of the system and without active exploration. The method selects a merging spot candidate by using the state value learned with pAC. We evaluate our method using real traffic data. Our experiments show that pAC achieves 92\% success rate to merge into a freeway, which is comparable to human decision making.
\end{abstract}

%------------------------------------------------------------------------------------------
\section{Introduction}
\label{sec:intro}
%------------------------------------------------------------------------------------------
Highly automated vehicles are growing in popularity \cite{GoogleCar,ziegler2014making,aeberhard15,okumura2016challenges} after DARPA2007 Urban Challenge\cite{darpa2007}. Merging in congested traffic is one of the tricky challenges toward fully automated driving. The difficulty is caused by three issues: how to model ambient vehicles, how to decide a merging spot and how to merge into the spot.

Multipolicy decision making (MPDM) has been developed in \cite{Galceran2017}. The approach decides based on multiple candidates with the scores of each policy according to user-defined cost function by using the forward simulation. Modeling ambient vehicles in congested traffic is needed for the forward simulation which is very difficult especially in congested traffic.

%Okumura et al.~\cite{okumura2016challenges} developed a method to merge into a roundabout with support vector machine. The method learns whether the ego-vehicle enter to the roundabout or not. The approach needs training labels. This approach also needs to prepare the each policy in advance.

%Bojarski et al.~\cite{bojarski2016end} developed end-to-end learning to merge into a roundabout. The method learns the policy with deep learning neural network.
Reinforcement learning is a powerful framework to learn policy without prior knowledge of environment by using active exploration. The score in MPDM can be calculated without forward simulation by using cumulative expected reward called state value as the score. The reinforcement learning, however, needs to explore the environment to optimize policy. An exhaustive exploration of the state and action spaces is impossible because it would be unacceptable for an autonomous vehicle to try maneuvers which lead to a crash or even a near-miss, leaving unexplored much of the state and action spaces.

Nishi et al.~(\citeyear{nishi2017}) developed a reinforcement learning method called {\em passive Actor Critic} (pAC), to optimize policy using data collected under passive dynamics and knowing dynamic model of ego-vehicle instead of such explorations. They showed that the method could learn improved policies in the highway merging domain with simulation and real traffic dataset. However in their experiment, the merging spot was decided in advance.

We develop a method for the freeway merging maneuver in congested traffic based on multi-policy decision making with pAC. The policy to merge into each candidate of merging spots is learned with pAC from collected data and ego-vehicle dynamics model. The method can choose a policy without forward simulation using instead the state-value function estimated by pAC.
%-----------------------------------------------------------------------------------------
\section{Preliminaries}
\label{sec:preliminaries}
%-----------------------------------------------------------------------------------------

We focus on a discrete-time system with a real-valued state $\x \in \mathbf{R}^n $ and control input $\mathbf{u} \in \mathbf{R}^m$, whose stochastic dynamics is defined as follows:
%{\fontsize{9.0pt}{10.0pt}\selectfont
\begin{align}
 \xkk  & = \xk + A(\xk)\dt + B\uk\dt + \diag\left(\boldsymbol{\sigma}\right)\boldsymbol{\Delta\omega},
 \label{eq:dynamics}
\end{align}
%}
where $\boldsymbol{\Delta\omega}$ is differential Brownian motion simulated by a Gaussian $\mathcal{N}(\mathbf{0},\mathbf{I}\dt)$, where $\mathbf{I}$ is the identity matrix. $A(\xk)$, $B\mathbf{\uk}$ and $\boldsymbol{\sigma} \in \mathbf{R}^n$ denote the {\em passive} dynamics, {\em control} dynamics due to action, and the transition noise level, respectively ($B$ is an input-gain matrix). $\Delta t$ is a step size of time and $k$ denotes a time index. System dynamics structured in this way are quite general: for example, models of many mechanical systems conform to these dynamics.

L-MDP~\cite{todorov2009efficient} is a subclass of MDPs defined by a tuple, $\langle \mathcal{X},\mathcal{U},\mathcal{P},\mathcal{R}\rangle$, where $\mathcal{X} \subseteq \mathbf{R}^n$ and $\mathcal{U} \subseteq \mathbf{R}^m$ are continuous state and action spaces. $\mathcal{P}\coloneqq \{p(\mathbf{y}|\mathbf{x},\mathbf{u}) \>|\> \mathbf{x},\mathbf{y} \in \mathcal{X}, \mathbf{u}\in \mathcal{U} \}$ is a state transition model due to action, which is structured as in Eq.~\ref{eq:dynamics}, and $\mathcal{R} \coloneqq \{r(\mathbf{x},\mathbf{u})\>|\> \mathbf{x}\in \mathcal{X}, \mathbf{u}\in \mathcal{U}\}$ is an immediate cost function with respect to state $\mathbf{x}$ and action $\mathbf{u}$. A control policy $\mathbf{u}=\pi(\mathbf{x})$ is a function that maps a state $\mathbf x$ to an action $\mathbf u$.
The goal is to find a policy that minimizes the following average expected cost: $\Vavg \coloneqq \lim_{n\rightarrow\infty}\frac{1}{n}\E\left[\sum_{k=0}^{n-1}r(\xk,\pi(\xk))\right]$.

Grondman et al.~(\citeyear{grondman2012survey}) notes that the Bellman equation for MDPs can be rewritten using the value function $V(\x)$ called {\em V-value}, state-action value function $Q(\x,\mathbf{u})$ called {\em Q-value}, and average value $V_{avg}$ under an policy.
\begin{align}
\Vavg + Q_{k} &= r_{k} + \E_{p(\xkk|\xu)} [ V_{k+1}].
\label{eq:Bellman_rev}
\end{align}
As we may expect, $V_{k} = \min_{\uu\in\mathcal{U}}Q_{k}$. $\E_{p(\xkk|\xk)}[\cdot]$ denotes expectation over a probability distribution of state transition under the passive dynamics. Here and elsewhere, subscript $k$ denotes values at time step $k$.

An L-MDP defines the cost of an action (control cost) to be the amount of additive stochasticity:
\begin{align}
r(\xu) \coloneqq q(\xk)\dt + KL(p(\xx)||p(\xxu)).
\label{eq:totalcost}
\end{align}
Here, $q(\mathbf{x})\geq 0 $ is the state-cost function; $KL(\cdot||\cdot)$ is the Kullback-Leibler (KL) divergence; $p(\xx)$ models the {\em passive} dynamics while $p(\xxu)$ represents the {\em active} or control dynamics of the system. L-MDPs further add a condition on the dynamics as shown below.
\begin{align*}
p(\xx) = 0 \Rightarrow \forall\uk~ p(\xxu) = 0.
\end{align*}
This condition ensures that no action introduces new transitions that are not achievable under passive dynamics. In other words, actions are seen as simply contributing to the passive dynamics. The stochastic dynamical system represented by Eq.~\ref{eq:dynamics} satisfies this assumption naturally because the dynamic is Gaussian. However, systems that are deterministic under passive dynamics remain so under active dynamics. This condition is easily met in robotic systems.

The standard Bellman equation for MDPs can then be recast in L-MDPs to be a linearized differential equation for exponentially transformed value function of Eq.~\ref{eq:Bellman_rev} (hereafter referred to as the linearized Bellman equation)~\cite{todorov2009eigenfunction}:
\begin{align}
\Zavg\Zk & = \expq~\E_{p(\xkk|\xk)}[\Zkk] \label{eq:definition-of-z},
\end{align}
where $\Zk \coloneqq e^{-V_k}$ and  $\Zavg \coloneqq e^{-\Vavg}$. Here, $\Zk$ and $Z_{avg}$ are an exponentially transformed value function called {\em Z-value} and the average cost under an optimal policy, respectively.

Because the passive and control dynamics with the Brownian noise are Gaussian, the KL divergence between these dynamics becomes
\begin{align}
KL(p(\xx)&||p(\xxu)) = 0.5\uk ^\top \Rinv \uk \dt,
\label{eq:rho_}
\end{align}
where $\Rinv \coloneqq B^\top ( \diag(\boldsymbol{\sigma}) \diag(\boldsymbol{\sigma})^\top )^{-1}B$. Then, the optimal control policy for L-MDPs can we derived as,
\begin{align}
\pi(\xk) & = -\R B^\top \frac{\partial V_k}{\partial \xk}.
\label{eq:optimraw}
\end{align}

%---------------------------------------------------------------------------------------
\section{Multi-policy decision making with pAC}
\label{sec:pAC}
%---------------------------------------------------------------------------------------
We present a novel MPDM algorithm with pAC. MPDM determines control input by selecting a policy in multiple candidates with the scores of each policy. While prior MPDM algorithm requires forward simulation to score each policy candidates, our algorithm scores the candidates without the simulation by using instead state value estimated with pAC. We provide details on the policy selection and pAC below.

\subsection{MPDM policy selection with pAC}
The MPDM policy selection returns the best policy based on a set of policy candidates and observed current state. Algorithm \ref{alg:MPDM} shows the policy selection algorithm. The algorithm sets $\Pi$ of available policies. Score $c$ of each candidate, which is calculated using state value estimated with pAC, is added to the set of scores $C$. Finally, the optimal policy associated with the minimum score is returned as the best policy.
\begin{algorithm}[!ht]
\caption{MPDM policy selection with pAC}\label{alg:MPDM}
\small
\begin{algorithmic}
\STATE Set $\Pi$ of available policies.
\STATE $C \leftarrow \emptyset$
\FOR {$\pi_i \in \Pi$}
\STATE Set current state: $\xk \leftarrow x$.
\STATE Calculate score of a policy $\pi_i$ learned by pAC: $c_i \leftarrow \hat{V}_i(\xk)$
\STATE $C \leftarrow C \cup c_i $
\ENDFOR
\STATE Choose the best policy $\pi^* \in \Pi$ : $\pi^* \leftarrow \arg\min_{\pi_i \in \Pi} C$
\RETURN $\pi^*$
\end{algorithmic}
\end{algorithm}

\subsection{Passive Actor-Critic for L-MDP}
We introduce an actor-critic method for continuous L-MDP, which we label as {\em passive actor-critic} (pAC). \trim{Figure~\ref{fig:overview} shows a detailed schematic of pAC.} While the actor-critic method usually operates using samples collected actively in the environment~\cite{konda1999actor}, pAC finds a converged policy without exploration. Instead, it uses samples of passive state transitions and a known control dynamics model. pAC follows the usual two-step schema of actor-critic: a state evaluation step (critic), and a policy improvement step (actor):
\begin{enumerate}[leftmargin=*, itemsep=0pt, topsep=0pt]
\item {\bf Critic}: Estimate the Z-value and the average cost from the linearized Bellman equation using samples under passive dynamics;
\item {\bf Actor}: Improve a control policy by optimizing the Bellman equation given the known control dynamics model, and the Z-value and cost from the critic.
\end{enumerate}
We provide details about these two steps below.

%~~~~~~~~~~~~~~~~~~~~~~~~~~~~~~~~~~~~~~~~~~~~~~~~~~~~~~~~~~~~~~~~~~~~~~~
\noindent{\bf Estimation by Critic using Linearized Bellman}
%~~~~~~~~~~~~~~~~~~~~~~~~~~~~~~~~~~~~~~~~~~~~~~~~~~~~~~~~~~~~~~~~~~~~~~~

The critic step of pAC estimates Z-value and the average cost by minimizing the least-square error between the true Z-value and estimated one denoted by $\hat{Z}$.
\begin{align}
\min_{\paramz,\hZavg}\frac{1}{2}\int _{\x}\Big( \hZavg\hat{Z}(\x;\paramz) -& \Zavg Z(\x)\Big)^2 d\x, \label{eq:ls_error}\\
{\bf s.t.} \, \int _{\x} \hat{Z}(\x;\paramz) d\x = C,&\,\, \forall {\mathbf x} \,\, 0 < \hat{Z}(\x;\paramz) \leq  \frac{1}{\hZavg}, \nonumber
\end{align}
where $\paramz$ is a parameter vector of the approximation and $C$ is a constant value used to avoid convergence to the trivial solution $\hat{Z}(\x;\paramz)= 0$ for all $\x$. The second constraint comes from $\forall {\x},\, Z(\x) \coloneqq e^{-V(\x)} > 0$ and $\forall {\mathbf x},\, q({\mathbf x})\geq 0$. The latter implies that $V + \Vavg > 0$, with $\Zavg Z(x) := e^{-(V+ \Vavg)}$ which is less than 1.

We minimize the least-square error in Eq.~\ref{eq:ls_error}, $\hZavg\Zkt - \Zavg\Zk$, with TD-learning. The latter minimizes TD error instead of the least-square error that requires the true $Z(\x)$ and $\Zavg$, which are not available. The TD error denoted as $e^{i}_k$ for linearized Bellman equation is defined using a sample $(\xk,\xkk)$ of passive dynamics as, $e^{i}_k \coloneqq \hZavg^{i}\Zkt^{i} - e^{-q_k}\Zkkt^{i}$, where the superscript $i$ denotes the iteration.
$\hZavg$ is updated using the gradient as follows:
\begin{align}
& \hZavg^{i+1} = \hZavg^{i} - \alpha_1^{i}\frac{\partial \left ( e^{i}_k \right)^2}{\partial \hZavg} = \hZavg^{i} - 2\alpha_1^{i} e^{i}_k \Zkt^{i},\label{eq:update_zavg}
\end{align}
where $\alpha^{i}_1$ is the learning rate, which may adapt with iterations.

In this work, we approximate the Z-value function using a neural network (NN). The parameters $\paramz$ are updated with the following gradient based on backpropagation:~\footnote{$e^{-\rm{tanh}(x)}$ or $e^{-\rm{softplus}(x)}$ is used as an activation function of the output layer to satisfy the constraint $\hat{Z}\geq 0$. The constraint $\int _{\x} \hat{Z}(\x;\paramz) d\x = C$  is ignored in practice because convergence to  $\forall{\x},\, \hat{Z}(\x;\paramz)=0$ is rare. $\min{([1,e^{-q_k}\Zkkt^{i}])}$ is used instead of $e^{-q_k}\Zkkt^{i}$ to satisfy $\hat{Z} \leq 1/\Zavg$ in Eq.~\ref{eq:ls_error}.}
\begin{align}
& \frac{\partial}{\partial \paramz^{i}}\left (\hZavg\Zkt^{i} - \Zavg\Zk  \right)^2 \approx 2e^{i}_k \hZavg^i\frac{\partial \Zkt^{i}}{\partial \paramz^{i}}, \label{eq:update_omega}
\end{align}
where $e_k^i$ is the TD error as defined previously.
\trim{$\nabla_{\paramz^{i}}\Zkkt^i$ is a partial derivative operator with respect to $\paramz^{i}$ on $\Zkkt^i$.}

%~~~~~~~~~~~~~~~~~~~~~~~~~~~~~~~~~~~~~~~~~~~~~~~~~~~~~~~~~~~~~~~~~~~~~~~
\noindent{\bf Actor Improvement using Standard Bellman}
%~~~~~~~~~~~~~~~~~~~~~~~~~~~~~~~~~~~~~~~~~~~~~~~~~~~~~~~~~~~~~~~~~~~~~~~

The actor improves a policy by computing $\R$ (Eq.~\ref{eq:rho_}) using the estimated Z-values from the critic because we do not assume knowledge of noise level  $\boldsymbol{\sigma}$. It is estimated by minimizing the least-square error between the V-value and the state-action Q-value:
\begin{align*}
\min_{\R} \frac{1}{2}\int _{\x}\left( \hat{Q}(\x,\hat{\uu}(\x)) - V(\x) \right)^2 d\x,
\end{align*}
where \trim{$M$ is the number of samples and a subscript $j$ is a sample index.} $V$ is the true V-value and $\hat{Q}$ is the estimated Q-value under the estimated action  $\hat{\uu}(\x)$. Notice from Eq.~\ref{eq:optimraw} that a value for $S$ results in a policy as $B$ is known. Thus, we seek the $S$ that yields the optimal policy by minimizing the least-square error because the Q-value equals V-value iff $\hat{\uu}$ is maximizing.

Analogously to the critic, we minimize the least-square error given above, $\hat{Q}^{i}_k - V_k$, with TD-learning. To formulate the TD error for the standard Bellman update, let $\xkk$ be a sample at the next time step given state $\xk$ under passive dynamics, and let $\tilde{\x}_{k+1} \coloneqq \xkk + B\huk\dt$ be the next state using control dynamics. Rearranging terms of the Bellman update given in Eq.~\ref{eq:Bellman_rev}, the TD error $d^{i}_k$ becomes
 \begin{align*}
 d^{i}_k \coloneqq r(\xk,\huk) + \hat{V}^{i}(\tilde{\x}_{k+1}) -\hVavg -\Vkt^{i}.
 \end{align*}
 We may use Eqs.~\ref{eq:totalcost} and~\ref{eq:rho_} to replace the reward function,
\begin{align*}
 r(\xk,\huk) &= (q_k + 0.5\huk^{\top}(\hR^{i})^{-1}\huk)\dt,\\
 %& = \qk + 0.5\hdVk^{i\top}B_k\hR^{i}B_k^{\top}\hdVk^{i}\dt + \hat{V}^{i}(\tilde{\x}_{k+1}) -\hVavg -\Vkt^{i}
 &= \qk + 0.5\frac{\partial \hat{V}_k^{i}}{\partial \xk}^{\top}B\hR^{i}B^{\top}\frac{\partial \hat{V}_k^{i}}{\partial \xk}\dt.
\end{align*}
The last step is obtained by noting that $\huk^{i} = -\hR^{i} B^{\top}\frac{\partial \hat{V}_k^{i}}{\partial \xk}$, where $\hR$ denotes estimated $\R$ in Eq. \ref{eq:optimraw}.
\trim{Here, $\hR^i$, $\Vkt^i$ and $\hdVk^{i}$ are the estimated $S$, the V-value and partial derivative of the V-value with respect to $\x$ at $\xk$ at iteration $i$, respectively. }

The estimated V-value and its derivative is calculated by utilizing the approximate Z-value function from the critic. $\hR$ is updated based on standard stochastic gradient descent using the TD error,
\begin{align*}
& \hR^{i+1} = \hR^{i} - \beta^i \frac{\partial}{\partial \hR^{i}}\left ( \hat{Q}^{i}_k - V_k \right)^2 \approx \hR^{i} - 2\beta^i d^{i}_k \frac{\partial d^{i}_k}{\partial \hR^{i}},
\end{align*}
where $\beta$ is the learning rate. The actor mitigates the impact of error from estimated Z-value by minimizing the approximated least-square error between V- and Q-values under the learned policy.
\trim{
The gradient of the TD error $d_k^i$ is used for updating the weight $\hR$ next, $\hR^{i+1} = \hR^{i} - \beta^i d^{i}_k \nabla_{\hR^{i}}d^{i}_k$ where $\beta^{i}$ is a learning rate. As the final step, the improved policy is obtained by using this updated estimate $\hR^{i+1}$ and derivative of updated V-value in Eq.~\ref{eq:optimraw}.}

%------------------------------------------------------------------------------
\noindent{\bf　Algorithm}
%------------------------------------------------------------------------------

\begin{algorithm}[!ht]
\caption{passive Actor Critic}\label{alg:pAC}
\small
\begin{algorithmic}
\STATE Initialize parameters $\hZavg^0, \paramz^0, \hR^0, \alpha_1^0, \beta^0$
\FOR {Iteration $i=1$ to $N$}
\STATE Sample set of a state, a next state and state cost $(\xk, \xkk, q_k)$ from dataset randomly
\BSTATE {\emph critic}:
\STATE $ e^{i}_k  \gets \hZavg^{i}\Zkt^{i} - \exp{(-q_k)}\Zkkt^{i}$
\STATE Update $\paramz^{i+1}$ with $2e^{i}_k \hZavg^i\frac{\partial \Zkt^{i}}{\partial \paramz^{i}}$
\STATE $\hZavg^{i+1} \gets \hZavg^{i} - 2\alpha_1^{i} e^{i}_k\Zkt^{i}$
\BSTATE \emph{actor}:
\STATE $\hat{V}_k^i \gets -\ln{\Zkt^i}$, $\hat{V}_{k+1}^i \gets -\ln{\Zkkt^i}$, $\hVavg^i \gets -\ln{\hZavg^i}$
\STATE $d^i_k \gets q_k + 0.5\frac{\partial \hat{V}_k^{i}}{\partial \xk}^{\top}B\hR^{i}B^{\top}\frac{\partial \hat{V}_k^{i}}{\partial \xk}\dt + \hat{V}_{k+1}^{i} -\hVavg^i -\Vkt^{i}$
\STATE $\hR^{i+1} \gets \hR^{i} - 2\beta^i d^{i}_k \frac{\partial d^{i}_k}{\partial \hR^{i}}$
\ENDFOR
\end{algorithmic}
\end{algorithm}

We show a pseudo code of pAC in Algorithm~\ref{alg:pAC}. $Z(\x)$ and $\Zavg$ are estimated in the critic with samples, and $\R$ is done in the critic with samples, estimated $\hat{Z}(\x)$ and $\hZavg$.
In the critic, feedback from the actor is not needed (unlike actor critic methods for MDPs) because the Z-value is approximated with samples from passive dynamics only. We emphasize that the actor and critic steps do not use the functions $A$ and $\boldsymbol{\sigma}$ but does indeed rely on $B$, of Eq.~\ref{eq:dynamics}. As such, the updates use a sample $({\mathbf x}_k,{\mathbf x}_{k+1})$ of the passive dynamics, and the state cost $q_k$.

%---------------------------------------------------------------------------------------
\section{Freeway merging based on MPDM}
\label{sec:freeway-merge}
%---------------------------------------------------------------------------------------

We present a freeway merging algorithm based on MPDM with pAC. The algorithm learns a policy and a state value function to merge into a predetermined spot with pAC from collected dataset in advance. The algorithm determines a merging spot from a set of candidates and control input with the learned model when an autonomous vehicle is driving on a merging ramp. We provide details on learning the policy and merging based on the learned policy below.

\subsection{Learning a merging policy in a predetermined spot}
\begin{figure}[!ht]
\centering
\includegraphics[width=1.0\hsize]{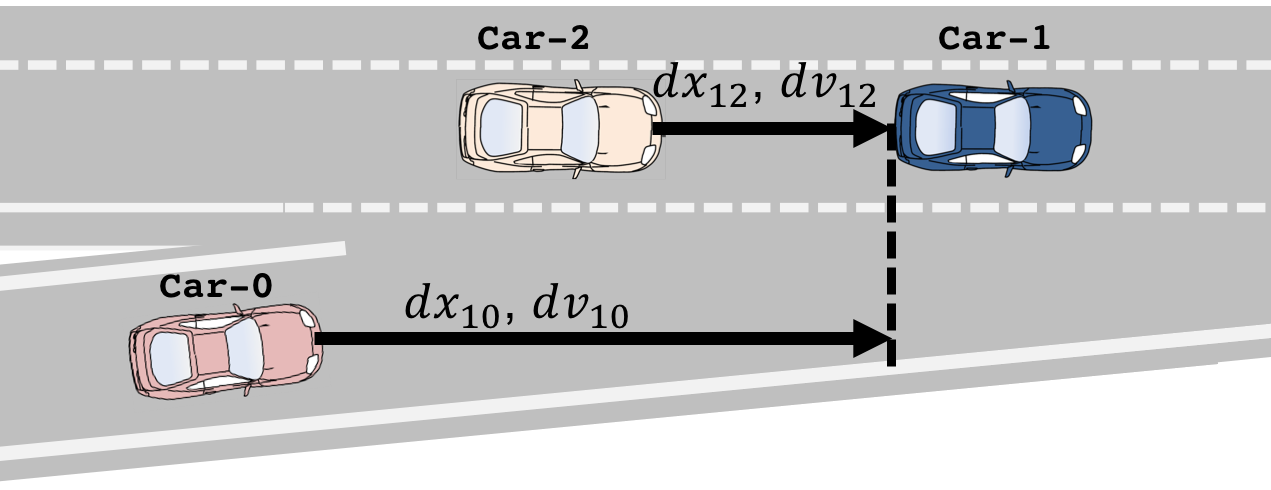}
\vskip -2mm
\caption{\small The 3-car system for merging. Merging vehicle (Car-0) should ideally merge midway between the following vehicle (Car-2) and leading vehicle (Car-1). $dx_{10}$ and $dv_{10}$ denote Car-0's relative position and velocity from Car-1.}
\label{fig:A-B-C}
\end{figure}

The algorithm learns a policy in a predetermined spot based on pAC with the following state space, action space and state cost function. We refer the reader to Fig.~\ref{fig:A-B-C} for our notation in the four-dimensional state space. Here, ${\mathbf x} = [dx_{12}, dv_{12}, dx_{10}, dv_{10}]^\top$ where $dx_{ij}$ and $dv_{ij}$ denote the horizontal signed distance and relative velocity between cars $i$ and $j\in [0,1,2]$. The action space is one-dimensional (acceleration). The state cost is designed to motivate Car-0 to merge midway between Car-1 and Car-2 with the same velocity as Car-2:
\begin{align*}
q({\mathbf x}) &= k_1  - k_1\exp \left( -k_2 \left(1-\frac{2dx_{10}}{dx_{10}}\right)^2 - k_3 dv_{12}^2\right),\\
& k_1=1,\,k_2 = 10,\,k_3=10\,\, {\rm if\,\,} dx_{12}<dx_{10}<0,\\
& k_1=10,\,k_2 = 10,\,k_3=0\,\, {\rm otherwise},
\end{align*}
where $k_1$, $k_2$ and $k_3$ are weights for the state cost.

\subsection{Freeway merging based on the learned policy}

\begin{figure}[!ht]
\centering
\includegraphics[width=1.0\hsize]{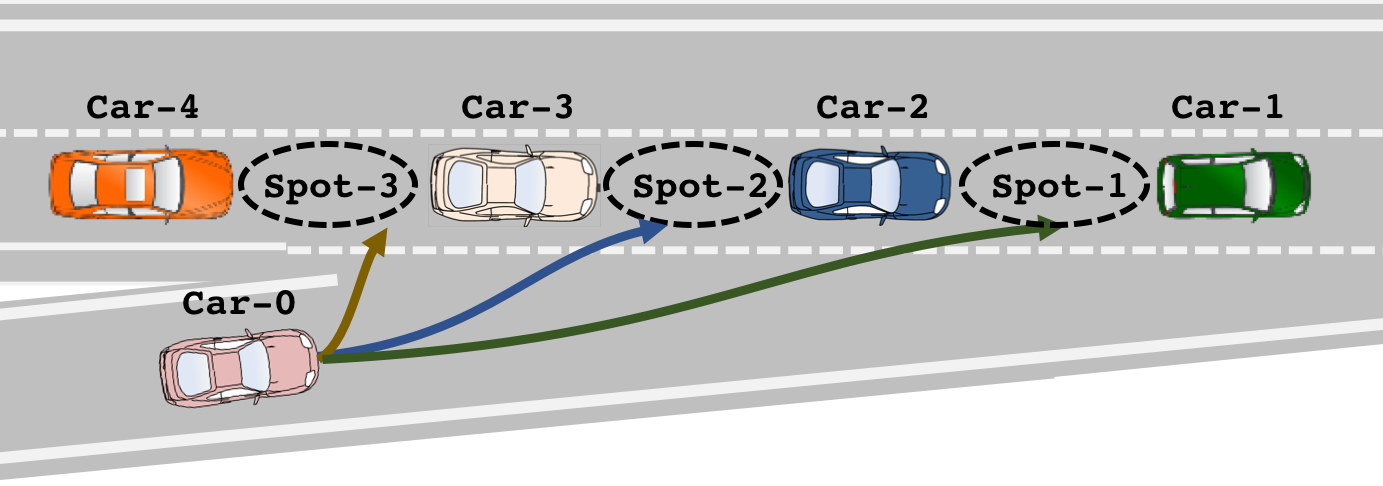}
\vskip -2mm
\caption{\small A typical freeway merging situation. There are three mergeable spot candidates: Spot-1, 2 and 3. The merging vehicle needs to determine a spot from a set of the candidates and control input to merge.}\label{fig:5-car-system}
\end{figure}
We present how to merge onto a freeway based on the learned policy. Figure \ref{fig:5-car-system} shows a typical freeway merging situation. In this situation, there are three possible mergeable spots: Spot-1, 2 and 3. First, the algorithm finds three mergeable spot candidates: Spot-1, 2 and 3. Three 3-car systems associated with the each spot are then extracted: \{Car-0, Car-1, Car-2\}, \{Car-0, Car-2, Car-3\} and \{Car-0, Car-3, Car-4\}.

 The best policy is selected with Algorithm \ref{alg:MPDM}. The state value function learned to merge into a predetermined spot can be used to calculate the scores of any spot candidates because the MDP is the same and only states are different between these candidates.

%--------------------------------------------------------------------------------
\section{Experiment on real-world traffic}
\label{sec:experiments}
%--------------------------------------------------------------------------------
We evaluate our algorithm with NGSIM data set recorded real-world traffic on freeway.

%~~~~~~~~~~~~~~~~~~~~~~~~~~~~~~~~~~~~~~~~~~~~~~~~~~~~~~~~~~~~~~~~~~~
\subsection{Problem settings}
%~~~~~~~~~~~~~~~~~~~~~~~~~~~~~~~~~~~~~~~~~~~~~~~~~~~~~~~~~~~~~~~~~~~
%\noindent {\bf Freeway merging in 5-car system} We introduce 5-car system to evaluate our method. The merging vehicle needs to merge into a spot in three candidates: C-1, C-2 and C-3 {\bf DP: could you clarify which spots, e.g., any place behind the following vehicle, ahead of it, or what?  It's hard to understand from the figure, and I don't think I understand the math describing the spot selection.}.

\begin{figure}[!ht]
\centering
\includegraphics[width=1.0\hsize]{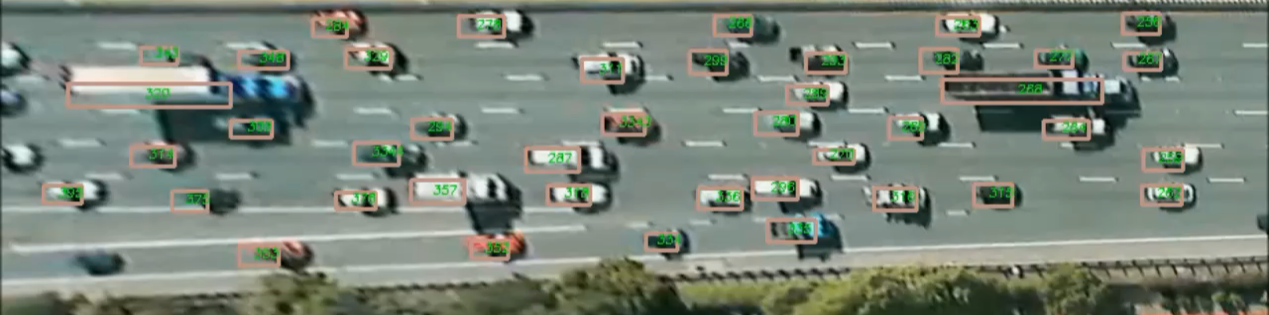}
\vskip -2mm
\caption{\small A snapshot of vehicles tracked in NGSIM. The bounding boxes denote the positions of tracked vehicles (their numbers are NGSIM indices).}\label{fig:ngsim-dataset}
	\end{figure}

The NGSIM data set contains vehicle trajectory data recorded by cameras mounted on top of a building on eastbound Interstate-80 in the San Francisco Bay area in Emeryville, CA for 45 minutes. It is segmented into three 15-minute periods around the evening rush hour \cite{ngsim}. In these periods, one can observe a transition from non-congested to moderately congested to full congestion. Vehicle trajectories were extracted using a vehicle tracking method from collected videos \cite{kovvali2007video}. We extracted 5-car system (Fig.~\ref{fig:5-car-system}) representing 637 freeway merging events, when human drivers always merged into Spot-2, and applied a Kalman smoother to mitigate vehicle tracking noise.　We calculated next states $\xkk$ under passive dynamics by subtracting state change caused by actions:$\xkk = \x^{D}_{k+1} - B^{\top}\uk^{D}$, where $\x^{D}_{k+1}$ and $\uk^{D}$ are the next state and the action recorded in the data set respectively.

%~~~~~~~~~~~~~~~~~~~~~~~~~~~~~~~~~~~~~~~~~~~~~~~~~~~~~~~~~~~~~~~~~~~~~~~~~~~
\subsection{Results}
\label{sec:ngsim}
%~~~~~~~~~~~~~~~~~~~~~~~~~~~~~~~~~~~~~~~~~~~~~~~~~~~~~~~~~~~~~~~~~~~~~~~~~~~

\begin{figure*}[!t]
\centering
\includegraphics[width=1.0\hsize]{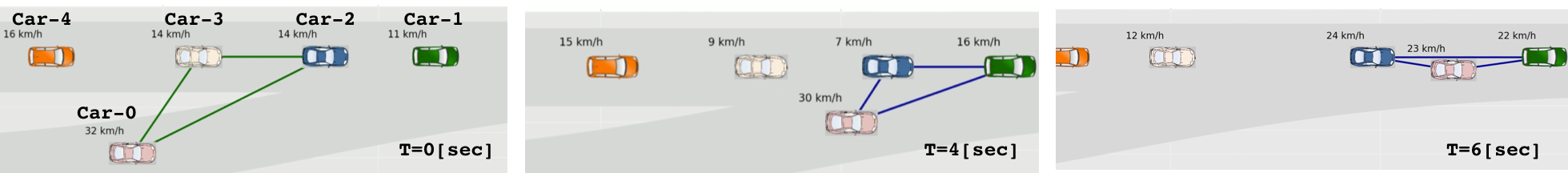}
\vskip -2mm
\caption{\small An example of freeway merging behavior with our algorithm. Triangles denote the 3-car system with the merging vehicle. {\bf (Left)} The vehicle tried to merge behind Car-2. {\bf (Center)}  The vehicle switched to merge behind Car-1 because it is easier to merge into the spot than others according to the scores calculated with MPDM. The switching would be reasonable because the vehicle would have needed hard braking if the vehicle had merged behind Car-2 due to large velocity gap between Car-0 and 2. {\bf (Right)} The vehicle is succeeding to merge behind Car-1. \trim{\bf DP: succeeding because the merging isn't completed yet, at least as shown in this figure.} }\label{fig:result-time-seriese}
\end{figure*}

Figure \ref{fig:result-time-seriese} shows an example of freeway merging behavior with our algorithm. The merging vehicle tried to merge behind Car-2 at T=0 seconds. The vehicle then switched to merge behind Car-1 and succeeded merging into the spot. The switching would be reasonable because the vehicle would have needed hard braking if the vehicle had merged behind Car-2 due to large velocity gap between Car-0 and Car-2.

\begin{figure*}[!t]
\centering
	\begin{tabular}{cc}
    	\begin{minipage}[t]{0.48\hsize}
        	\includegraphics[width=1.0\hsize]{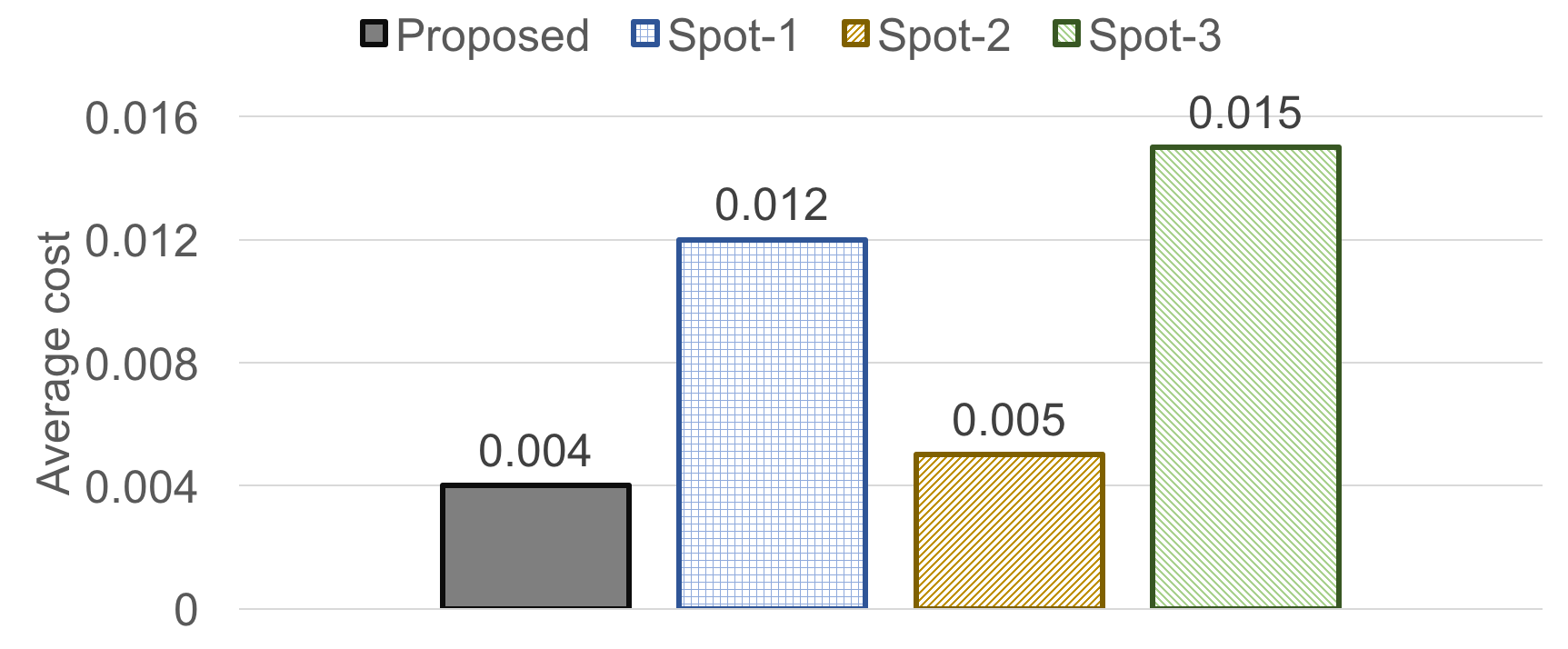}
    	\end{minipage}
    	\begin{minipage}[t]{0.48\hsize}
        	\includegraphics[width=1.0\hsize]{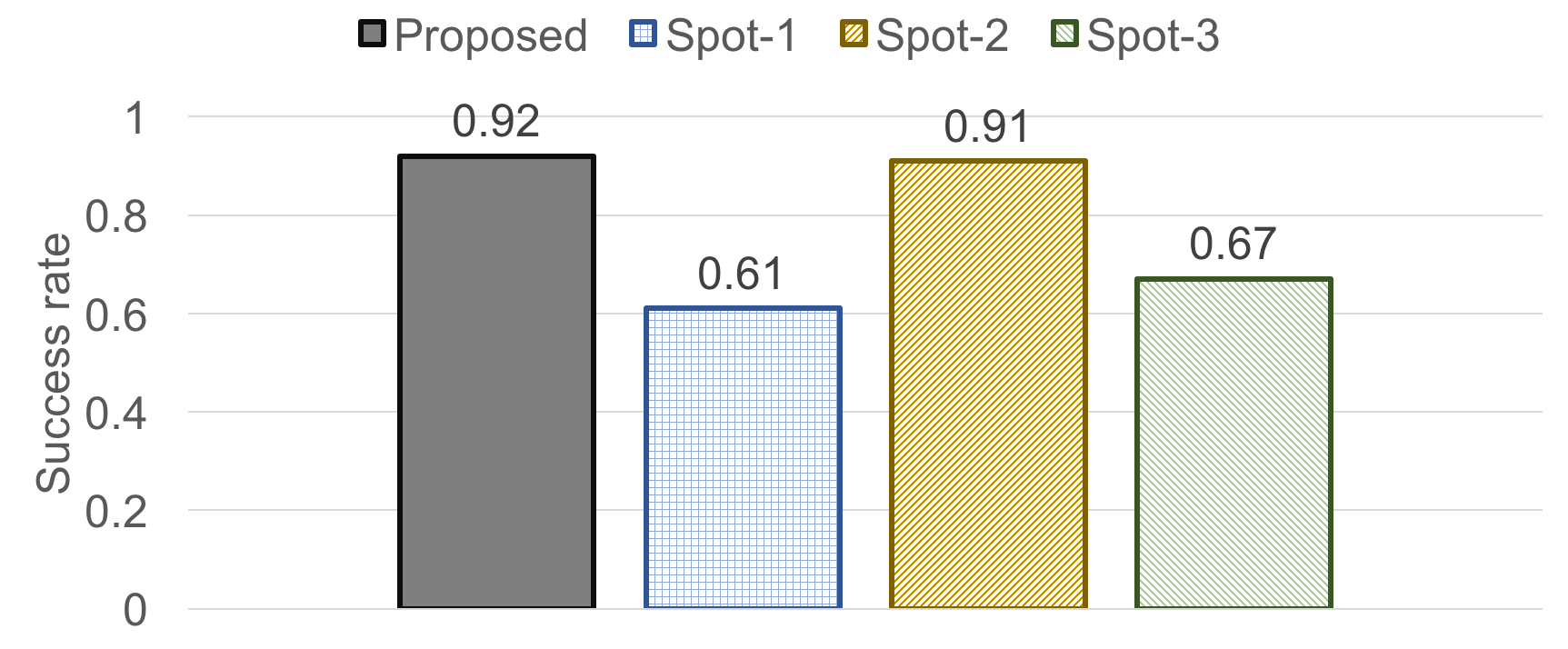}
    	\end{minipage}
	\end{tabular}
	\caption{\small {\bf (Left)} Average costs of each policy. The policy with the proposed method(gray) is comparable to a policy to merge into Spot-2 (orange) chosen by human drivers. {(\bf Right)} Success rate of each policy. Our method achieved a 92\% success rate. The performance is comparable to a success rate of the policy to merge always into Spot-2.}\label{fig:results}
\end{figure*}

We compared a policy based on the proposed method to three policy to merge into each of the predetermined spots: Spot-1, 2 and 3. The left chart of Fig. \ref{fig:results} shows average costs of each policy to merge into Spot-1, 2 and 3. Our method achieved comparable average cost to the policy to merge into Spot-2 chosen by human drivers, and outperform other policies. The right chart of Fig. \ref{fig:results} shows a success rate of freeway merging into each spot. The result also shows that our method can select proper merging spots because the performance is better than that of the policy to merge into Spot-2.  Our method fails occasionally when a following vehicle strongly accelerates to reduce the gap or a leading vehicle strongly decelerates.  \trim{\bf DP: Can we report something about our failures?  Would be useful, as people could argue that we are only slightly better than the Spot-2 policy.}

%-----------------------------------------------------------------------------
\section{Concluding Remarks}
\label{sec:conclusion}
%-----------------------------------------------------------------------------

We presented a novel MPDM method with pAC. The method needs no forward simulation because it uses estimated state value with pAC unlike prior MPDM methods. This feature would be preferable when we install our MPDM method onto autonomous vehicles because smaller computational resources would be needed to calculate the score with estimated state value function than for any approach employing forward simulations.

We also illustrated our MPDM on a freeway merging task. The algorithm determines a merging spot and a policy to merge into a predetermined spot with the learned model and a state value function after extracting all possible 3-car systems.

We evaluated the method using freeway merging domain on real traffic data. Our method achieved a 92\% success rate, which is comparable to merging success if the spot is selected by human drivers. We are interested in improving performance of our method further, as well as exploring additional challenges, e.g., deciding when to enter roundabouts.
%---------------------------------------------------------------------------

\bibliographystyle{icml2017}

\end{document}